%% file: acl2019.tex
\documentclass[11pt,a4paper]{article}
\usepackage[hyperref]{acl2019}
\usepackage{times}
\usepackage{latexsym}
\usepackage{rotating}
\usepackage{booktabs}
\usepackage{url}
\usepackage{graphicx}
\usepackage{todonotes}
\usepackage{natbib}
\usepackage{color}
\usepackage[utf8]{inputenc} 
\usepackage{verbatim}
\usepackage[toc,page]{appendix}

\aclfinalcopy 
\def\aclpaperid{***} 

\newcommand*{\affaddr}[1]{#1} 
\newcommand*{\affmark}[1][*]{\textsuperscript{#1}}

\title{Automatic Domain Adaptation Outperforms Manual
  Domain Adaptation
  for Predicting Financial Outcomes}

\author{{\centering
Marina Sedinkina{\rm\affmark[1]} ~~
Nikolas Breitkopf{\rm\affmark[2]} ~~ 
Hinrich Sch\"{u}tze{\rm\affmark[1]}} \vspace{.15cm}
\\
\affaddr{\affmark[1]Center for Information \& Language Processing, LMU Munich} \\
\affaddr{\affmark[2]Institute for Finance \& Banking, LMU Munich}\\
\affaddr{\texttt{sedinkina@cis.uni-muenchen.de}}}

\date{}

\def\mathindent{\mbox{\hspace{0.5cm}}}

\newcounter{notecounter}
\newcommand{\enotesoff}{\long\gdef\enote##1##2{}}
\newcommand{\enoteson}{\long\gdef\enote##1##2{{
\stepcounter{notecounter}
{\large\bf
\hspace{1cm}\arabic{notecounter} $<<<$ ##1: ##2
$>>>$\hspace{1cm}}}}}
\enoteson
\enotesoff

\def\tabref#1{Table~\ref{tab:#1}}
\def\tablabel#1{\label{tab:#1}\label{p:#1}}

\def\secref#1{\S\ref{sec:#1}}
\def\seclabel#1{\label{sec:#1}\label{p:#1}}
\def\eqref#1{Eq.~\ref{eqn:#1}}

\def\dnrm#1{\mbox{$_{\hbox{\scriptsize #1}}$}}

\long\def\eat#1{}

\begin{document}
\maketitle

\begin{abstract}
In this paper, we  automatically create
sentiment dictionaries for predicting
financial outcomes. We compare three approaches:
(i)
manual adaptation of
the domain-general dictionary H4N, (ii)
automatic adaptation of
H4N and (iii)
a combination consisting of first manual, then automatic adaptation.
In our experiments, we demonstrate that the
automatically adapted 
sentiment dictionary outperforms the previous state of the
art
in predicting the financial outcomes
\emph{excess return} and \emph{volatility}.
In particular, automatic adaptation performs better than
manual adaptation. In our analysis, we find
that annotation based on 
\emph{an expert's a priori belief}
about a word's meaning can be incorrect -- annotation should
be performed based on the word's \emph{contexts in the
  target domain} instead.
\end{abstract}

\section{Introduction}
Since 1934, the U.S. Securities and Exchange Commission
(SEC) mandates that
public companies
disclose information in form of public filings
to ensure that  adequate information is available to
investors. One such filing is the 10-K,
the company's annual report. It 
contains
financial statements and
information about business strategy, 
risk factors and legal issues. For this reason, 10-Ks are
an important  source of information in
the field of finance and accounting.

A common method
employed by 
finance and accounting
researchers
is
to evaluate the ``tone''  of a text based
on the Harvard
Psychosociological Dictionary, specifically, on the
Harvard-IV-4 TagNeg (H4N) word
list.\footnote{\url{http://www.wjh.harvard.edu/~inquirer}}
However, as its name suggests, this dictionary is from a
domain that is different from finance, so many words (e.g.,
``liability'', ``tax'') that are labeled as negative in H4N
are in fact not negative in finance.

In a pioneering study,
\citet{donald} manually
reclassified the words in H4N for the financial domain.
They applied the resulting
dictionaries\footnote{\url{https://sraf.nd.edu/textual-analysis/resources}} to 10-Ks and
predicted financial variables such as
excess return and
volatility.
We will refer to the sentiment dictionaries created
by \citet{donald} as L\&M.

In this work, we
also create sentiment dictionaries for the finance domain,
but we adapt them from the domain-general H4N dictionary
\emph{automatically}.
We first learn word embeddings from a corpus of 10-Ks and then
reclassify them -- using SVMs trained on H4N labels -- as negative vs. non-negative.
We refer to  the resulting domain-adapted dictionary as
H4N$\dnrm{RE}$. 

In our experiments, we demonstrate that the
automatically adapted financial
sentiment dictionary H4N$\dnrm{RE}$ performs better at 
predicting 
excess return and volatility than dictionaries of 
\citet{donald} and \citet{theil}.

We make the following contributions.
\textbf{(i)}  We demonstrate that 
automatic domain adaptation
performs better at 
predicting financial outcomes
than previous work based on manual domain adaptation.
\textbf{(ii)} We perform an analysis of the
  differences between the classifications of
L\&M and those of
  our sentiment dictionary H4N$\dnrm{RE}$ that sheds light on the
  superior performance of H4N$\dnrm{RE}$.
For example, H4N$\dnrm{RE}$ is much smaller than L\&M, consisting
mostly of frequent words, suggesting H4N$\dnrm{RE}$ is more robust and
less prone to overfitting.
\textbf{(iii)}
In a further detailed analysis, we
investigate words
classified by L\&M as 
\textit{negative}, \textit{litigious} and
\textit{uncertain} that our embedding classifier classifies otherwise; and 
common (i.e., non-negative) words from H4N that 
L\&M did not include in the categories
\textit{negative}, \textit{litigious} and
\textit{uncertain}, but that our embedding classifier
classifies as belonging to these classes. Our analysis
suggests that manual adaptation of dictionaries is
error-prone if annotators are not given access to
corpus contexts.

Our paper primarily addresses a finance application. In
empirical finance, a correct sentiment classification
decision is not sufficient -- the decision must also be
\emph{interpretable} and \emph{statistically sound}.  That is why we use
ordinary least squares (OLS) -- an
established method  in empirical finance -- and sentiment dictionaries. Models based
on sentiment dictionaries are transparent and interpretable:
by looking at the dictionary words occurring in a document
we can trace the classification decision back to the original
data and, e.g., understand the cause of a classification
error. OLS is a well-understood statistical method that
allows the analysis of significance, effect size and
dependence between predictor variables, inter alia.

While we focus on finance here, three important lessons of
our work also
apply to many other domains. (1) An increasing number of
applications require interpretable 
analysis; e.g., the European Union
mandates that systems
used for sensitive
applications provide  explanations of decisions.
Decisions based on a solid statistical foundation are more
likely to be trusted than those by black boxes.
(2) 
Many NLP applications are domain-specific and require
domain-specific
resources including lexicons. Should such lexicons be built
manually from scratch or adapted from generic lexicons? We
provide evidence that automatic adaptation works better.
(3) Words often have specific meanings in a domain and this
increases the risk that a word is misjudged
if only the generic meaning is present to the
annotator. This seems to be the primary reason for the
problems of manual lexicons in our experiments. Thus, if
manual lexicon creation is the only option, then it is
important to present words in context, not in isolation, so
that the domain-specific sense can be recognized.

\section{Related Work}
In \textbf{empirical finance}, researchers have exploited
various text resources, e.g., news \citep{Kazemian16},
microblogs \citep{semeval17}, twitter \citep{zamani2017using} and
company disclosures \citep{Nopp15, Kogan09}. Deep learning has been used for
learning document representations \citep{Ding15, Akhtar17}.
However, the methodology of empirical finance requires
interpretable results. Thus, a common approach is
to define
features for statistical models like Ordinary Least Squares
\citep{Lee_onthe, Rekabsaz17}.  Frequently, lexicons like 
H4N TagNeg\footnote{\url{http://www.wjh.harvard.edu/~inquirer/}}
\citep{tetlock} are used. It includes a total of 85,221 words,
4188 of which are labeled negative. The remaining words are
labeled ``common'', i.e., non-negative.
\citet{donald} argue that many
words from H4N have a specialized meaning when appearing in an
annual report. For instance, domain-general negative words such as ``tax'', ``cost'',
   ``liability'' and ``depreciation''
-- which predominate
in 10-Ks -- do not typically have negative sentiment in 10-Ks. So
\citet{donald} constructed subjective financial dictionaries manually, by examining all
words that appear in at least 5\% of 10-Ks
and classifying them based on their
assessment of most likely usage. More recently, other finance-specific lexicons
were created \cite{Wang13}. Building on L\&M,
\citet{tsai} and \citet{theil}
show that the L\&M dictionaries can be further improved by
adding most similar neighbors to words
manually labeled by L\&M.

\textbf{Seed-based methods}
generalize a set of seeds based on corpus (e.g.,
distributional) evidence.
Models use syntactic  patterns \citep{Hatzivassiloglou,
  Widdows}, cooccurrence  \citep{Turney, Igo}
or label propagation on lexical graphs derived from 
cooccurrence \cite{Velikovich10,Huang}.

\textbf{Supervised methods}
start with a larger training set, not just a few seeds
\citep{Mohammad}. 
Distributed word representations
\citep{Tang14,Amir,Vo,Rothe} are beneficial in this approach. For instance, \citet{Tang14} incorporate in word embeddings a
document-level sentiment signal. 
\citet{Wang17} also integrate
document and word levels. 
\citet{Hamilton16} learn domain-specific word
embeddings and derive word lists specific to  domains,
including the finance domain.

\textbf{Dictionary-based approaches} \citep{Takamura,
  BaccianellaES10, Vicente14} use hand-curated lexical
resources -- often WordNet \citep{fellbaum98wordnet} -- for
constructing lexicons.  \citet{Hamilton16} argue that
dictionary-based approaches generate better results due to
the quality of hand-curated resources. We compare two ways
of using a hand-curated resource in this work -- a
general-domain resource that is automatically adapted to the
specific domain vs.\  a resource that is manually
created for the specific domain -- and show that automatic
domain adaptation performs better.

Apart from domain adaptation work on dictionaries, many
other approaches to \textbf{generic domain adaptation} have
been proposed. Most of this work adopts the classical domain
adaptation scenario: there is a large labeled training set
available in the source domain and an amount of
labeled target data that is insufficient
for training a high-performing model on its own
\citep{blitzer2006domain,chelba2006adaptation,daume2009frustratingly,pan2010cross,glorot2011domain,chen2012marginalized}. More
recently, the idea of domain-adversarial training was
introduced for the same scenario \citep{ganin2016domain}.
In contrast to this work, we do not transfer any
parameters or model structures from source to
target. Instead, we use labels from the source domain and
train new models from scratch based on these labels: first
embedding vectors, then a classifier that is trained on
source domain labels and finally a regression model that is
trained on the classification decisions of the classifier.
This approach is feasible in our problem setting because the
divergence between source and target sentiment labels is
relatively minor, so that training target embeddings with
source labels gives good results.

The motivation for this different setup is that
our work primarily addresses a finance application where
explainability is of high importance.
For this reason, we  use a model based on sentiment
dictionaries that allows us to provide explanations of the model's
decisions and predictions.

\enote{hs}{

  A large part of the work has focused on
adapting a model trained on the source domain for use in the
target domain. For example,  use
the source model as a prior on the weights for a second
model, trained on the target data. In contrast,
 optimizes source and target
parameters jointly, not sequentially.  They augment the
feature space of both the source and target data and use the
result as input to a standard learning
algorithm. Concurrently, multiple methods proposed learning
a common feature representation that is meaningful across
both domains.  So,  adapt
Structural Correspondence Learning (SCL) to identify
correspondences among features from different
domains.  use a spectral feature
alignment (SFA) algorithm to find a robust representation
for cross-domain data.  More recently, neural network
representations have become increasingly studied
.  In these
approaches, the classifier is learned in a separate step
using the feature representations.  In contrast,
domain-adversarial learning approach of
\citet{ganin2016domain} performs feature learning, domain
adaptation and classifier learning jointly.
}

\eat{
Our work differs in the sense that we use a general-domain lexicon and
adapt it to a specialized domain. We argue that domain-specific
dictionaries can be constructed automatically without having to rely
on a manually compiled domain-specific dictionary. 
}

\section{Methodology}
\seclabel{compfin}
\subsection{Empirical finance methodology}
In this paper, we adopt Ordinary Least Squares (OLS),
a common research method in empirical finance: a
dependent variable of interest (e.g., excess return,
volatility) is predicted based on a linear combination of a
set of explanatory variables.

The main focus of this paper
is to investigate text-based explanatory variables: we would
like to know to what extent a text variable such as
occurrence of negative words in a 10-K can predict a
financial variable like volatility.
Identifying the economic drivers of such a financial outcome
is
of central interest in the field of finance.
Some of these determinants may be correlated with sentiment.
To understand the role of sentiment in explaining financial
variables
we therefore need to isolate the \emph{complementary
  information} of our text variables.
This is achieved by including in our
regressions -- as control variables -- a standard set of
financial explanatory variables such as firm size and 
book-to-market ratio. These control variables are added as additional explanatory variables in the regression specification besides the textual sentiment variables.
This experimental setup allows us to assess the added benefit of text-based
variables in a realistic empirical finance scenario.

The approach is
motivated by previous studies in the finance literature (e.g.,
\citet{donald}), which show that 
characteristics
of financial firms can
explain variation in excess returns and volatility. By including these
control variables in the regression we are able to determine whether
sentiment factors have incremental explanatory power beyond the
already established financial factors. Since the inclusion of these
control variables is not primarily driven by the assumption that firms
with different characteristics use different language, our approach
differs from other NLP studies, such as \citet{hovy2015demographic}, who accounts for non-textual characteristics by training group-specific embeddings.

Each text variable we use is based on a dictionary. Its
value for a 10-K is the proportion of tokens in the 10-K that
are members of the dictionary. For example, if the 10-K is
5000 tokens long and 50 of those tokens are contained in the
L\&M uncertainty dictionary, then the value of the L\&M uncertainty
text variable for this 10-K is 0.01.

In the type of analysis of stock market data we conduct, there are two general forms of
dependence in the residuals of a regression, which arise from the panel structure of our data set  where a single firm is repeatedly observed over time and
  multiple firms are observed at the same point in time.
\emph{Firm effect:}
Time-series
dependence assumes that the residuals of \emph{a given firm} are
correlated \emph{across years}.
\emph{Time effect:}
Cross-sectional dependence assumes
that the residuals of \emph{a given year} are correlated \emph{across
  different firms}.
These properties violate the i.i.d.\ assumption of
residuals in standard  OLS.
We therefore model data
with both  firm and time effects and run
a \emph{two-way robust cluster regression}, i.e.,
an OLS
regression with standard errors that are clustered on two
dimensions \cite{gelbach2009robust}, the dimensions of firm
and time.\footnote{\citet{donald} use 
  the method of
\citet{fama1973risk} instead.
This method assumes that the yearly
estimates of the coefficient are independent of each
other. However, this is not true when there is a firm effect.}
We apply this regression-based methodology to test the explanatory power of financial dictionaries with regard to two dependent variables:
excess return and volatility. 
This approach allows us to compare the explanatory power of different sentiment dictionaries and in the process test the hypothesis that negative sentiment is associated with subsequently lower stock returns and higher volatility.
We now introduce the regression specifications for these tests.

\subsubsection{Excess return}
\seclabel{excess-sec}
The dependent variable 
excess return is defined as the firm's
buy-and-hold stock return minus the
value-weighted buy-and-hold market index return during the
4-day event window
starting on the \mbox{10-K} filing date,
computed from prices by 
the Center for Research in
Security Prices (CRSP)\footnote{\url{http://www.crsp.com}}
(both expressed as a percentage).
In addition to the independent text variables (see
\secref{exp} for details), we include the following
financial control variables.
(i) Firm size:  the log of the
book value of total assets.
(ii) Alpha of a Fama-French regression
\citep{FAMA1933}
calculated
from days \mbox{[-252 -6]};\footnote{[-252 -6] is the notation for
  the 252 days prior to the filing date
  with the last 5 days prior to the filing date excluded.}
this represents the ``abnormal'' return of the asset, i.e.,
the part of the return not due to common risk factors like
 market and firm size.
(iii)
Book-to-market ratio:   the log of
the book value of equity divided by the market value of
equity. (iv) Share turnover:  the volume of shares
traded in days [-252 -6] divided by shares outstanding on
the filing date. (v) 
Earnings surprise,
  computed by IBES from Thomson Reuters;\footnote {\url{http://www.thomsonreuters.com}}
this variable captures
whether the reported financial performance was better or
worse than expected by
  financial analysts.\footnote{Our setup
largely mirrors, but
  is
  not identical to the one used by  \citet{donald}
  because not all data they  used are publicly
  available and because we use a larger time window
  (1994-2013) compared to theirs (1994-2008).}

\subsubsection{Volatility}
The dependent variable 
volatility is defined as
the post-filing root-mean-square error (RMSE) of a Fama-French regression calculated
from days [6 252].
The RMSE captures the idiosyncratic component of the total volatility of the firm, since it
picks up the stock price variation that cannot be explained by fluctuations of the common risk factors of the Fama-French model.
The RMSE is therefore a measure of the financial uncertainty of the firm.
In addition to the independent text variables (see \secref{exp} for details), we include the following
financial control variables.
(i)
Pre-filing RMSE and (ii) pre-filing alpha of a Fama-French regression
calculated
from days [-252 -6];
these characterize the financial uncertainty and abnormal return
of the firm in the past
(see \secref{excess-sec} for alpha and first sentence of this section
for RMSE).
(iii) Filing abnormal return; the value of the
buy-and-hold return in trading days [0 3] minus the
buy-and-hold return of the market index.
(iv) Firm size and (v) book-to-market ratio (the same
as in \secref{excess-sec}).
(vi) Calendar year dummies
and Fama-French 48-industry dummies to allow for time  and
industry fixed effects.\footnote{We do not include in the
regression a Nasdaq dummy variable indicating whether the firm
is traded on Nasdaq. Since  Nasdaq 
mainly lists tech companies, the Nasdaq effect is
already captured by  industry dummies.}

\begin{table}
\begin{center}
\small
  \begin{tabular}{l|r}
    dictionary & size\\\hline\hline
neg$\dnrm{lm}$  &  2355\\ 
unc$\dnrm{lm}$  &  297\\
lit$\dnrm{lm}$  &  903\\ \hline  
neg$\dnrm{ADD}$ & 2340\\ 
unc$\dnrm{ADD}$ & 240 \\ 
lit$\dnrm{ADD}$  &  984  \\\hline
neg$\dnrm{RE}$  & 1205 \\ 
unc$\dnrm{RE}$ &96   \\ 
lit$\dnrm{RE}$   & 208  \\  \hline
H4N$\dnrm{ORG}$  &4188\\ 
H4N$\dnrm{RE}$   &338 
 \end{tabular}

\end{center}
\caption{\tablabel{overview-add}Number of words per dictionary}
\end{table}

\subsection{NLP methodology}
\seclabel{nlpmethod}
There are two
main questions we want to answer:

\mathindent
  \textbf{Q1.} Is a manually domain-adapted or an automatically
    domain-adapted
    dictionary a more effective predictor of financial
    outcomes?

\mathindent    \textbf{Q2.} L\&M adapted H4N for the financial domain and
    showed that this manually adapted dictionary is more
    effective than H4N for prediction. Can we further
    improve L\&M's manual adaptation by automatic domain
    adaptation?

The general methodology we employ for domain adaptation is
based on word embeddings. We train CBOW word2vec
\citep{micolov-13} word embeddings on a corpus of 10-Ks for
all words of H4N that occur in the corpus -- see
\secref{exp} for details.
We consider two adaptations: ADD and RE. ADD is only used to
answer question Q2.

\textbf{ADD.}  For adapting the L\&M dictionary, we train an
SVM on an L\&M dictionary in which words are labeled +1 if
they are marked for the category by L\&M and labeled -1
otherwise (where the category is negative, uncertain or
litigious).  Each word is represented as its embedding.  We
then run the SVM on all H4N words that are not contained in the 
L\&M dictionary. We also ignore H4N words that we do not
have embeddings for because their frequency is below the
word2vec frequency threshold. Thus, we obtain an ADD dictionary which
is not a superset of the L\&M lexicon because it
includes only new additional words that are not part of the original
dictionary.

SVM scores
are converted into probabilities via logistic regression. We
define a confidence threshold $\theta$ -- we only want
to include words in the ADD dictionary  that are
reliable indicators of the category of interest. A word 
is added to the dictionary if its converted SVM score 
is greater
than $\theta$.

\textbf{RE.}  We train SVMs
as for ADD, but this time in a five-fold cross validation
setup. Again, SVM scores are converted into probabilities
via logistic regression.
A word $w$ becomes a member of the
adapted dictionary
if its converted SVM score of the SVM that was not
trained on the fold that contains $w$ is greater than
$\theta$.

To answer our first
question Q1: ``Is automatic or manual adaptation better?'',
we apply adaptation method RE to H4N
and compare the results to the L\&M dictionaries.

To answer our second
question Q2: ``Can manual adaptation be further improved by
automatic adaptation?'',
we apply adaptation methods RE and ADD to
the three dictionaries compiled by L\&M and compare results
for original  and adapted L\&M dictionaries: (i) negative (abbreviated as ``neg''), (ii) uncertain (abbreviated as ``unc''), (iii) litigious (abbreviated as ``lit'').
Our goals here are to
improve 
the
  in-domain
  L\&M dictionaries by relabeling them using adaptation method RE
  and to find new additional words using adaptation method ADD.

\tabref{overview-add} gives dictionary sizes.

\section{Experiments and results}
\seclabel{exp}
We downloaded 206,790 10-Ks for years 1994 to 2013 from the
SEC's database
EDGAR.\footnote{https://www.sec.gov/edgar.shtml} 
Table of contents, page numbers,
links and numeric tables are removed
in preprocessing
and
only the main body of the text is retained.
Documents are split into sections. Sections that are not
useful for textual analysis (e.g., boilerplate) are deleted.

To construct the final sample, we apply the 
filters defined by L\&M \citep{donald}: we require a match with CRSP's permanent
identifier PERMNO, the stock to be common equity, a stock
pre-filing price of greater than \$3, a positive
book-to-market, as well as CRSP's market capitalization and
stock return data available at least 60 trading days before
and after the filing date. We only keep firms traded on
Nasdaq, NYSE or AMEX and whose filings contain at least 2000
words.
This procedure results in a corpus of
60,432 10-Ks.
We  tokenize (using NLTK)
and lowercase this corpus and remove punctuation.

We use word2vec CBOW with hierarchical softmax to
learn word embeddings from the corpus.
We set the size of word vectors to
400 and run one training iteration; otherwise we use
word2vec's default
hyperparameters. SVMs are trained on word embeddings as
described in \secref{nlpmethod}.
We set the threshold $\theta$
to 0.8, so only words with converted SVM scores greater than
0.8 will be added to dictionaries.\footnote{We choose this
  threshold because the proportion of negative, litigious
  and uncertain words in 10-Ks for 0.8 is roughly the same as when using
  L\&M dictionaries.}

As described in \secref{compfin}, we compare manually
adapted and automatically adapted dictionaries (Q1) and
investigate whether automatic adaptation of manually adapted
dictionaries further improves performance (Q2).
Our experimental setup is Ordinary Least Squares (OLS), more specifically,
a two-way robust cluster regression for the time and
firm effects.
The dependent financial variable is excess return or
volatility. We include several independent financial
variables in the regression as well as one or more text variables.
The value of the text variable for a category  is
the proportion of tokens from the category that occur in a 10-K.

To assess the utility of a text variable for predicting a
financial outcome, we look at
significance and
the standardized regression coefficient (the product of
regression coefficient and standard deviation).
If a result is significant, then it is unlikely 
that the result is
due to chance. The standardized coefficient measures the
effect size, normalized for different value ranges of
variables.
It can be interpreted as the expected change in the dependent variable 
if the independent variable increases by one standard deviation.
The standardized coefficient allows
a fair comparison between a text variable that, on average,
has high values (many tokens per document) with one that, on
average, has low values (few tokens per document).

\begin{table}
\begin{center}
  \small
\begin{tabular}{l|llll}
 var &  coeff &  std coeff &  t &  $R^2$\\ \hline
 neg$\dnrm{lm}$  & -0.202**&-0.080 & \bf-2.56 & 1.02\\ 
 lit$\dnrm{lm}$   &-0.0291&-0.026& -0.83 &1.00\\ 
 unc$\dnrm{lm}$  &-0.215*&-0.064 &\bf -1.91 &1.01 \\ 
 H4N\dnrm{RE}  &-0.764***&\textit{-0.229}& \textbf{-3.04} &1.05\\ \hline
\multicolumn{5}{c}{*p $\le$ 0.05, **p $\le$ 0.01, ***p $\le$ 0.001}
 \end{tabular}
\end{center}
\caption{\tablabel{harvard-excess}
  Excess
  return regression results for L\&M dictionaries and reclassified H4N
  dictionary.
\textbf{For all tables in this paper, 
significant $t$
  values are bolded and best standard coefficients per category
  are in italics.}}
\end{table}

\begin{table}
\begin{center}
  \small
\begin{tabular}{l|llll}
 var &  coeff &  std coeff &  t &  $R^2$\\ \hline\hline
  H4N\dnrm{RE} & -0.88** &\textit{-0.264} & \textbf{-2.19}& 1.05\\ 
 neg$\dnrm{lm}$  & \phantom{-}0.062& \phantom{-}0.024& \phantom{-}0.48&\\ \hline
 H4N\dnrm{RE}&-0.757*** &-0.227& \textbf{-2.90}&1.05\\ 
 lit$\dnrm{lm}$   &-0.351&\textit{-0.315}&-0.013 &\\ \hline
  H4N\dnrm{RE} &-0.746*** &\textit{-0.223} &\textbf{-2.89} &1.05 \\ 
 unc$\dnrm{lm}$  &-0.45&-0.135 &-0.45 &\\ \hline\hline
\multicolumn{5}{c}{*p $\le$ 0.05, **p $\le$ 0.01, ***p $\le$ 0.001}
 \end{tabular}

\end{center}
\caption{\tablabel{multi-excess}Excess return regression results
  for multiple text variables. This table shows
  results for three regressions that combine H4N$\dnrm{RE}$
  with each of the three L\&M dictionaries.}
\end{table}

\subsection{Excess Return}
\tabref{harvard-excess}
gives regression results 
for excess return, comparing 
H4N$\dnrm{RE}$ (our automatic adaptation of the general
Harvard dictionary) with the three manually adapted L\&M dictionaries.
As expected the
coefficients are negatively signed --  10-Ks containing  a high
percentage of pessimistic words are associated with negative excess returns.

L\&M designed the dictionary 
neg$\dnrm{lm}$ specifically for measuring negative
information in a 10-K that may have a negative effect on
outcomes like excess return. So it is not surprising that 
neg$\dnrm{lm}$ is the best performing dictionary of the
three L\&M dictionaries: it has the highest standard
coefficient (\mbox{-0.080}) and the highest significance (-2.56).
unc$\dnrm{lm}$ performs slightly worse, but is also
significant.

However, when comparing the three L\&M dictionaries with
H4N$\dnrm{RE}$, the automatically adapted Harvard
dictionary, we see that H4N$\dnrm{RE}$ performs clearly
better: it is highly significant and its standard
coefficient is larger by a factor of more than 2 compared to
neg$\dnrm{lm}$.  This evidence suggests that the
automatically created H4N$\dnrm{RE}$ dictionary has a higher
explanatory power for excess returns than the manually
created L\&M dictionaries. This provides an initial answer
to question Q1: in this case, automatic adaptation beats
manual adaptation.

\def\mycolspace{0.125cm}
\begin{table}
\begin{center}
  \small
  \begin{tabular}{l|l@{\hspace{\mycolspace}}l@{\hspace{\mycolspace}}r@{\hspace{\mycolspace}}l}
 var &  coeff &  std coeff &  \multicolumn{1}{c}{$t$} &  $R^2$\\ \hline\hline
 neg$\dnrm{lm}$  & -0.202** &-0.080 & \bf-2.56 & 1.02\\ 
neg$\dnrm{spec}$  &\phantom{-}0.0102 &\phantom{-}0.0132&0.27 &1.00\\ 
 neg$\dnrm{RE}$  & -0.37*** &\emph{-0.111}&  \textbf{-2.96} & 1.03\\ 
 neg$\dnrm{ADD}$ & -0.033 &-0.0231 &  -1.03 & 1.00 \\ 
 neg$\dnrm{RE+ADD}$   & -0.08** &-0.072 & \bf -2.19 & 1.03\\\hline
 lit$\dnrm{lm}$   &-0.0291 &-0.026& -0.83 &1.00\\ 
 lit$\dnrm{RE}$  &-0.056 &\emph{-0.028}& -0.55 &1.00\\ 
 lit$\dnrm{ADD}$ &-0.0195 &-0.0156&-0.70 &1.00\\ 
 lit$\dnrm{RE+ADD}$    &-0.0163&-0.0211& -0.69 &1.00 \\\hline
 unc$\dnrm{lm}$  &-0.215* &-0.064& \bf -1.91 &1.01 \\ 
 unc$\dnrm{RE}$  &-0.377*** &\emph{-0.075}& \bf -2.77 &1.02\\ 
 unc$\dnrm{ADD}$ &\phantom{-}0.0217 &\phantom{-}0.0065& 0.21 &1.00\\ 
 unc$\dnrm{RE+ADD}$    &-0.0315 &-0.0157 &-0.45 &1.00\\\hline\hline
\multicolumn{5}{c}{*p $\le$ 0.05, **p $\le$ 0.01, ***p $\le$ 0.001}
 \end{tabular}
\end{center}
\caption{\label{add-ro} Excess return  regression results for
  L\&M, \textbf{RE} and
  \textbf{ADD} dictionaries}
\end{table}

\tabref{multi-excess} shows \textit{manual plus automatic}
experiments with \emph{multiple} text variables in one regression, in
particular, the combination of H4N\dnrm{RE} with each of the L\&M
dictionaries. We see that the explanatory power of
L\&M variables is lost after we additionally include 
H4N\dnrm{RE}  in a regression: all three L\&M
variables are not significant. In contrast, H4N\dnrm{RE}
continues to be significant in all experiments, with large
standard coefficients. More manual plus automatic
experiments can be found in
the appendix.
These experiments further confirm that automatic is better than manual adaptation.

Table \ref{add-ro} shows results for automatically adapting
the L\&M dictionaries.\footnote{Experiments with multiple text variables in one
regression (manual plus automatic experiments) are presented
in the appendix.}
The subscript ``RE+ADD'' refers to a dictionary
that merges RE and ADD; e.g., 
neg$\dnrm{RE+ADD}$ is the union of
neg$\dnrm{RE}$ and
neg$\dnrm{ADD}$.

We see that for each category (neg, lit and unc), the automatically
adapted dictionary performs better than the original
manually adapted dictionary; e.g., the standard coefficient
of
neg$\dnrm{RE}$ is -0.111, clearly better than that of
neg$\dnrm{lm}$  (-0.080). Results are significant for 
neg$\dnrm{RE}$ (-2.96) and
unc$\dnrm{RE}$ (-2.77). 
We also evaluate neg$\dnrm{spec}$, the negative word list of \citet{Hamilton16}. neg$\dnrm{spec}$ does not perform well: it is not significant.

These results provide a partial answer
to question Q2: for excess return, automatic adaptation
of L\&M's manually adapted dictionaries further improves
their performance.

\begin{table}
\begin{center}
  \small
\begin{tabular}{l|llll}
  var &  coeff &  std coeff &  \multicolumn{1}{c}{$t$} &  $R^2$\\ \hline\hline
neg$\dnrm{lm}$  &\phantom{-}0.118*** &\phantom{-}0.0472 &\phantom{-}\textbf{3.30} & 60.1\\ 
lit$\dnrm{lm}$  &-0.0081 &-0.0073&-0.62 & 60.0 \\ 
unc$\dnrm{lm}$  &\phantom{-}0.119* & \phantom{-}0.0356 &\phantom{-}\textbf{2.25} & 60.0 \\ 
H4N\dnrm{RE}  &\phantom{-}0.577*** &\textit{\phantom{-}0.173}& \phantom{-}\textbf{4.40} &60.3\\\hline\hline
\multicolumn{5}{c}{*p $\le$ 0.05, **p $\le$ 0.01, ***p $\le$ 0.001}
 \end{tabular}
\end{center}
\caption{\tablabel{harvard-vol} Volatility regression results  for L\&M dictionaries and reclassified H4N dictionary}
\end{table}

\begin{table}
\begin{center}
  \small
\begin{tabular}{l|llll}
 var &  coeff &  std coeff &  t &  $R^2$\\ \hline \hline
  H4N\dnrm{RE}  & \phantom{-}0.748*** & \textit{\phantom{-}0.224}& \phantom{-}\textbf{4.44} & 1.11\\ 
 neg$\dnrm{lm}$  & -0.096*& -0.038& -2.55& \\ \hline
 H4N\dnrm{RE}  &\phantom{-}0.642*** &\textit{\phantom{-}0.192}&\phantom{-}\textbf{4.28}&1.11\\
 lit$\dnrm{lm}$   &-0.041*&-0.037& -2.54 &\\ \hline
  H4N\dnrm{RE}  &\phantom{-}0.695*** &\phantom{-}0.208 &\phantom{-}\textbf{4.54} &1.11\\ 
unc$\dnrm{lm}$  &-0.931**&\textit{-0.279 }& -2.73 &\\ \hline \hline
\multicolumn{5}{c}{*p $\le$ 0.05, **p $\le$ 0.01, ***p $\le$ 0.001}
 \end{tabular}
\end{center}
\caption{\tablabel{multiple-vol} Volatility regression results for multiple text variables}
\end{table}

\subsection{Volatility}
\tabref{harvard-vol} compares H4N$\dnrm{RE}$ and L\&M regression results 
for volatility. Except for litigious, the coefficients are
positive, so
the greater the number of pessimistic words, the greater the volatility.

Results for neg$\dnrm{lm}$, unc$\dnrm{lm}$
and H4N$\dnrm{RE}$
are statistically
significant.  The best L\&M dictionary is again
neg$\dnrm{lm}$ with standard coefficient 0.0472 and
$t=3.30$. However,
H4N\dnrm{RE} has the highest explanatory value for volatility.
Its standard coefficient (0.173) 
is more than three times as large as that of neg$\dnrm{lm}$.

The higher effect size
demonstrates that
H4N\dnrm{RE} 
better explains
volatility than the L\&M dictionaries. Again, this indicates
-- answering question Q1 -- that
automatic outperforms manual adaptation.
\tabref{multiple-vol} confirms this. We see that for manual plus automatic experiments each
combination of H4N\dnrm{RE} with one of the L\&M
dictionaries provides significant results for H4N\dnrm{RE}.
In contrast, L\&M dictionaries become negatively signed
meaning that more uncertain words decrease volatility,
suggesting that they are not indicative of the true
relationship between volatility and negative tone in 10-Ks
in this regression setup.  Our results of additional manual plus
automatic experiments support this observation as well. See
the appendix for an illustration.

\begin{table}
\begin{center}
  \small
  \begin{tabular}{l|l@{\hspace{\mycolspace}}l@{\hspace{\mycolspace}}r@{\hspace{\mycolspace}}l}
 var &  coeff &  std coeff &  \multicolumn{1}{c}{$t$} &  $R^2$\\ \hline\hline
neg$\dnrm{lm}$  &\phantom{-}0.118*** &\phantom{-}0.0472 &\textbf{3.30} & 60.1\\ 
neg$\dnrm{spec}$  &-0.038 &-0.0494&-2.73 &60.1\\ 
neg$\dnrm{RE}$  &\phantom{-}0.219*** &\phantom{-}\emph{0.0657} &\textbf{3.57} & 60.1\\ 
neg$\dnrm{ADD}$  &\phantom{-}0.032*** &\phantom{-}0.0224& \textbf{4.06} & 60.0\\ 
neg$\dnrm{RE+ADD}$  &\phantom{-}0.038*** &\phantom{-}0.0342&\textbf{4.32}& 60.1\\ \hline
lit$\dnrm{lm}$  &-0.0081 &-0.0073 &-0.62 & 60.0 \\ 
lit$\dnrm{RE}$  & \phantom{-}0.0080 &\phantom{-}0.0040 & 0.20 &  60.0 \\ 
lit$\dnrm{ADD}$  & 0.028&\phantom{-}\textit{0.0224} & 1.07 &  60.0 \\ 
lit$\dnrm{RE+ADD}$  & 0.015 &\phantom{-}0.0195&0.81 &  60.0 \\\hline 

unc$\dnrm{lm}$  & \phantom{-}0.119* &\phantom{-}\emph{0.0356} &\textbf{2.25} &  60.0 \\ 
unc$\dnrm{spec}$  & -0.043 &-0.0344 & -1.56 &  60.0 \\ 
unc$\dnrm{RE}$  &\phantom{-}0.167* &\phantom{-}0.0334&\textbf{2.30} &  60.0\\ 
unc$\dnrm{ADD}$  &-0.013 &-0.0039& -0.17&  60.0\\ 
unc$\dnrm{RE+ADD}$  &\phantom{-}0.035&\phantom{-}0.0175 &0.68 & 60.0\\ \hline \hline
\multicolumn{5}{c}{*p $\le$ 0.05, **p $\le$ 0.01, ***p $\le$ 0.001}
 \end{tabular}
\end{center}
\caption{\label{volat} Volatility regression results  for L\&M, \textbf{RE} and
  \textbf{ADD} dictionaries}
\end{table}

Table \ref{volat} gives results for automatically adapting
the L\&M dictionaries.\footnote{Experiments with multiple text variables in one
regression (manual plus automatic experiments) are presented
in the appendix.} For neg,
the standard coefficient of neg$\dnrm{RE}$ is 0.0657, better
by about 40\% than 
neg$\dnrm{lm}$'s standard coefficient of 0.0472. 
neg$\dnrm{spec}$ does
not provide significant results and has the negative sign, i.e., an
increase of negative words decreases volatility. 
The lit dictionaries are not significant (neither L\&M nor adapted dictionaries).
For unc, 
unc$\dnrm{RE}$ performs worse than 
unc$\dnrm{lm}$, but only slightly by 0.0344 vs.\ 0.0356 for
the standard coefficients.
The overall best result is
neg$\dnrm{RE}$ (standard coefficient 0.0657).
Even though L\&M designed the unc$\dnrm{lm}$ dictionary
specifically for volatility, our results indicate that neg
dictionaries perform better than unc dictionaries, both for
L\&M dictionaries (neg$\dnrm{lm}$) and their automatic adaptations (e.g., neg$\dnrm{RE}$).

Table \ref{volat} also evaluates 
unc$\dnrm{spec}$, the uncertainty dictionary of
\citet{theil}. unc$\dnrm{spec}$ does not perform well: it is not
significant and the coefficient has the ``wrong'' sign.\footnote{\citet{theil} define volatility for the time period
  [6 28] whereas our definition is 
 [6 252], based on \citep{donald}. Larger time windows allow
  more reliable estimates and account for the fact that
  information
  disclosures can influence volatility for
  long  periods
  \citep{zhao}.}

The main finding supported by Table \ref{volat} is
that
the best automatic
adaptation of an L\&M dictionary gives rise to more explanatory power
than the best L\&M dictionary, i.e.,
neg$\dnrm{RE}$ performs better than neg$\dnrm{lm}$.
This again confirms our answer to Q2: we can further improve manual adaptation by automatic domain adaptation.

\begin{table}
  \begin{center}
    \small
\begin{tabular}{l|l}
ADD\dnrm{neg}  & missing, diminishment, disabling, overuse\\ 
ADD\dnrm{unc}  & reevaluate, swings, expectation, estimate\\  
ADD\dnrm{lit}  & lender, assignors, trustee, insurers \\ \hline
RE\dnrm{neg} & confusion, unlawful, convicted, breach\\
RE\dnrm{unc}  & variability, fluctuation, variations, variation\\ 
RE\dnrm{lit}  & courts, crossclaim, conciliation, abeyance\\ \hline
H4N$\dnrm{RE}$
& compromise, issues, problems, impair, hurt
 \end{tabular}

\end{center}
\caption{\label{new-words}Word classification examples from automatically adapted dictionaries}
\end{table}

\section{Analysis and discussion}
\subsection{Qualitative Analysis}
\label{qu:a}
\seclabel{qualana}
Our dictionaries outperform L\&M.
In this section, we perform a qualitative analysis to
determine the reasons for this discrepancy in performance.

Table \ref{new-words} shows words
from automatically adapted dictionaries.
Recall that the
\textbf{ADD}
method adds words that L\&M classified as nonrelevant for a category.
 So words like ``missing'' (neg), ``reevaluate'' (unc) and
 ``assignors'' (lit) were classified as relevant terms
 and seem to connote
negativity,  uncertainty and
litigiousness, respectively, in financial
contexts.

In L\&M's classification scheme, a word can
be part of several different categories.
For instance, L\&M label  ``unlawful'', ``convicted'' and
``breach''
both as
litigious and as negative.
When applying our RE method, these words were only
classified as negative, not as litigious.
Similarly, L\&M label ``confusion'' as negative and
uncertain, but automatic RE adaptation labels it only negative.
This indicates that there is strong distributional evidence in the
corpus for the category negativity, but weaker
distributional evidence for litigious and uncertain. For our
application, only ``negative'' litigious/uncertain words are
of interest -- ``acquittal'' (positive litigious) and ``suspense''
(positive uncertain) are examples of positive words that may
not help in predicting financial variables. This could
explain why the negative category fares better in our
adaptation than the other two.

An interesting case study for RE is  ``abeyance''.
L\&M classify it as uncertain, automatic adaptation as litigious.
Even though ``abeyance'' has a domain-general uncertain sense
(``something that is waiting to be acted
upon''), it is mostly used in legal contexts in 10-Ks:
``held in abeyance'', ``appeal in
  abeyance''.
  The nearest neighbors of
``abeyance'' in embedding space are also litigious words:
``stayed'', ``hearings'', ``mediation''.

H4N$\dnrm{RE}$ contains 74 words that are  ``common'' in
H4N. Examples include
``compromise'', 
``serious'' and ``god''.
The nearest neighbors of
``compromise'' in the 10-K embedding space
 are the negative terms ``misappropriate'',
``breaches'', ``jeopardize''.
In a general-domain embedding
space,\footnote{\url{https://code.google.com/archive/p/word2vec/}} the
nearest neighbors of ``compromise'' include ``negotiated settlement'', 
``accord'' and ``modus vivendi''.
This example suggests that ``compromise'' is used in  10-Ks in 
negative contexts and in the general domain in positive contexts.
 This also illustrates the importance of domain-specific word
embeddings that  capture domain-specific information.

Another interesting example is the word ``god''; it is
frequently used in 10-Ks in the phrase ``act of
  God''.
Its nearest
neighbors in the 10-K embedding space  are ``terrorism'' and
   ``war''.
This example clearly demonstrates that annotators are likely
to make mistakes when they annotate words for sentiment
without seeing their contexts. Most annotators would
annotate ``god'' as positive, but when presented with the
typical context in 10-Ks (``act of God''), they would be
able to correctly classify it.

We 
conclude that manual
annotation
of words without context based on the prior belief an
annotator has about word meanings is error-prone.
Our automatic adaptation
is performed based on
the word's contexts in the target domain and therefore not
susceptible to this type of error.

\def\confspace{0.075cm}

\begin{table}
  \begin{center}
    {\small
\begin{tabular}{l||r@{\hspace{\confspace}}r@{\hspace{\confspace}}r|r@{\hspace{\confspace}}r@{\hspace{\confspace}}r|r@{\hspace{\confspace}}r@{\hspace{\confspace}}r|r@{\hspace{\confspace}}r@{\hspace{\confspace}}r}
&{\rotatebox{90}{neg$\dnrm{lm}$}}&{\rotatebox{90}{lit$\dnrm{lm}$}}&{\rotatebox{90}{unc$\dnrm{lm}$}}&{\rotatebox{90}{neg$\dnrm{ADD}$}}&{\rotatebox{90}{lit$\dnrm{ADD}$}}&{\rotatebox{90}{unc$\dnrm{ADD}$}}&{\rotatebox{90}{neg$\dnrm{RE}$}}&{\rotatebox{90}{lit$\dnrm{RE}$}}&{\rotatebox{90}{unc$\dnrm{RE}$}}&{\rotatebox{90}{H4N$\dnrm{neg}$}}&{\rotatebox{90}{H4N$\dnrm{cmn}$}}&{\rotatebox{90}{H4N$\dnrm{RE}$}}\\\hline\hline
neg$\dnrm{lm}$ &&7&2&0&0&0&49&2&0&48&52&12 \\
lit$\dnrm{lm}$ &17&&0&0&0&0&6&20&0&7&93&1 \\
unc$\dnrm{lm}$ &14&0&&0&0&0&18&2&30&16&84&2\\\hline
neg$\dnrm{ADD}$ &0&0&0&&0&0&0&0&0&18&82&2\\
lit$\dnrm{ADD}$ &0&0&0&0&&0&0&0&0&1&99&0\\
unc$\dnrm{ADD}$ &0&0&0&0&0&&0&0&0&3&97&0\\\hline
neg$\dnrm{RE}$ &95&5&4&0&0&0&&0&1&52&48&21\\
lit$\dnrm{RE}$ &18&86&2&0&0&0&0&&0&7&93&0\\
unc$\dnrm{RE}$ &11&2&92&0&0&0&10&0&&13&87&3\\\hline
H4N$\dnrm{neg}$&27&2&1&10&0&0&15&0&0&&0&6\\
H4N$\dnrm{cmn}$&2&1&0&2&1&0&1&0&0&0&&0\\
H4N$\dnrm{RE}$&79&2&2&17&0&0&74&0&1&78&22&
\end{tabular}}
\caption{\label{analysis-qu}
  Quantitative analysis of dictionaries.
For a row dictionary $d_r$ and a column dictionary $d_c$, a
cell
gives $|d_r \cap d_c|/|d_r|$ as a percentage.
 Diagonal entries (all equal to 100\%) omitted for space
 reasons. cmn = common}
\end{center}
\end{table}

\subsection{Quantitative Analysis}

Table \ref{analysis-qu} presents a quantitative analysis of
the distribution of words over dictionaries.  For a row
dictionary $d_r$ and a column dictionary $d_c$, a cell gives
$|d_r \cap d_c|/|d_r|$ as a percentage.  (Diagonal entries
are all equal to 100\% and are omitted for space reasons.)
For example, 49\% of the words in neg\dnrm{lm} are also
members of neg\dnrm{RE} (row ``neg\dnrm{lm}'', column
``neg\dnrm{RE}'').  This analysis allows us to obtain
insights into the relationship between different
dictionaries and into the relationship between the
categories negative, litigious and uncertain.

Looking at rows 
neg$\dnrm{lm}$,
lit$\dnrm{lm}$ and
unc$\dnrm{lm}$ first, we see how L\&M constructed their
dictionaries. 
neg$\dnrm{lm}$ words come from 
H4N$\dnrm{neg}$ and
H4N$\dnrm{cmn}$ in about equal proportions; i.e., many words
that are ``common'' in ordinary usage were classified as
negative by L\&M for financial text. Relatively few 
lit$\dnrm{lm}$ and
unc$\dnrm{lm}$ words are taken from
H4N$\dnrm{neg}$, most are from
H4N$\dnrm{cmn}$. Only 12\% of neg$\dnrm{lm}$ words were
automatically classified as negative in domain adaptation
and assigned to H4N$\dnrm{RE}$. This is a surprisingly low
number. Given that H4N$\dnrm{RE}$ performs better than 
neg$\dnrm{lm}$ in our experiments, this statistic casts
serious doubt on the ability of human annotators to
correctly classify words for the type of sentiment analysis
that is performed in empirical finance if the actual corpus
contexts of the words are not considered. We see two
types of failures in the human annotation. First, as
discussed in \secref{qualana}, words like ``god'' are
misclassified because the prevalent context in 10-Ks (``act
of God'') is not obvious to the annotator. Second,
the utility of a word is not only a function of its
sentiment, but also of the strength of this sentiment. Many
words in neg$\dnrm{lm}$ that were deemed neutral in
automatic adaptation are probably words that may be slightly
negative, but that do not contribute to explaining financial
variables like excess return. The strength of sentiment of a
word is
difficult to judge by human annotators.
Looking at the row H4N$\dnrm{RE}$, we see that most of its
words are taken from neg$\dnrm{lm}$ (79\%) and a few from
lit$\dnrm{lm}$ and unc$\dnrm{lm}$ (2\% each). We can
interpret this statistic as indicating that L\&M had high
recall (they found most of the reliable indicators), but low
precision (see the previous paragraph: only 12\% of their
negative words survive in H4N$\dnrm{RE}$). The distribution
of H4N$\dnrm{RE}$ words over H4N$\dnrm{neg}$
and H4N$\dnrm{cmn}$ is 78:22. This confirms the need for
domain adaptation: many general-domain common words are
negative in the financial domain.

We finally look at how dictionaries for negative, litigious
and uncertain overlap, separately for the L\&M, ADD and RE dictionaries.
lit$\dnrm{lm}$ and
unc$\dnrm{lm}$ have considerable overlap with neg$\dnrm{lm}$
(17\% and 14\%), but they do not overlap with each other.
The three ADD dictionaries --
neg$\dnrm{ADD}$,
lit$\dnrm{ADD}$ and
unc$\dnrm{ADD}$ -- do not overlap at all.
As for RE, 10\% of the
words of unc$\dnrm{RE}$ are also in 
neg$\dnrm{RE}$, otherwise there is no overlap between RE
dictionaries. Comparing the original L\&M dictionaries and
the automatically adapted ADD and RE dictionaries, we see
that the three categories -- negative, litigious and
uncertain -- are more clearly distinguished after
adaptation. L\&M dictionaries overlap more, ADD and RE
dictionaries overlap less.

\section{Conclusion}
In this paper, we  automatically created
sentiment dictionaries for predicting
financial outcomes. 
In our experiments, we demonstrated that the
automatically adapted 
sentiment dictionary H4N$\dnrm{RE}$ outperforms the previous state of the
art
in predicting the financial outcomes
excess return and volatility.
In particular, automatic adaptation performs better than
manual adaptation.
Our quantitative and qualitative study provided insight into the
semantics of the dictionaries.
We found
that annotation based on 
\emph{an expert's a priori belief}
about a word's meaning can be incorrect -- annotation should
be performed based on the word's \emph{contexts in the
  target domain} instead. 
In the future, we plan to investigate whether there are
changes over time that significantly
impact the linguistic characteristics of the data, in the
simplest case changes in the meaning of a word.  Another 
interesting topic for future research is
the comparison of domain adaptation based on our
domain-specific word embeddings vs.\ based on word embeddings trained on much larger corpora.

\section*{Acknowledgments}
We are grateful for the support of the European Research
Council for this work (ERC \#740516).

\bibliography{naaclhlt2019}
\bibliographystyle{acl_natbib}
\appendix
\clearpage
\section{Appendix}
\subsection{Excess return regression results for multiple text variables}
\input{appendix_tex/full_reg_4d_post_HA_nocontr}
\input{appendix_tex/full_reg_4d_post_RE_neg_nocontr}
\input{appendix_tex/full_reg_4d_post_RE_unc_nocontr}
\input{appendix_tex/full_reg_4d_post_RE_lit_nocontr}\clearpage

\subsection{Volatility regression results for multiple text variables}
\input{appendix_tex/vol_HA_nocontr}
\input{appendix_tex/vol_neg_nocontr}
\input{appendix_tex/vol_unc_nocontr}
\input{appendix_tex/vol_lit_nocontr}

\end{document}

%% file: appendix_tex/full_reg_4d_post_HA_nocontr.tex
\begin{table}[h]
\begin{center}
  \small
\begin{tabular}{l|llll}
 var &  coeff &  std coeff &  t &  $R^2$\\ \hline\hline
  H4N\dnrm{RE} & -0.88** &\textit{-0.264} & \textbf{-2.19}& 1.05\\ 
 neg$\dnrm{lm}$  & \phantom{-}0.062& \phantom{-}0.024& \phantom{-}0.48&\\ \hline

 H4N\dnrm{RE}&-0.739** &\textit{-0.221}& \textbf{-2.23}&1.05\\ 
 all$\dnrm{lm}$   &-0.008&-0.008&-0.21 & \\ \hline

 H4N\dnrm{RE}&-0.836** &\textit{-0.25}& \textbf{-2.15}&1.05\\ 
 neg\_unc$\dnrm{lm}$   &\phantom{-}0.027&\phantom{-}0.016&\phantom{-}0.28 & \\ \hline

 H4N\dnrm{RE}&-0.755** &\textit{-0.226}& \textbf{-2.56}&1.05\\ 
 neg\_lit$\dnrm{lm}$   &-0.003&-0.004&-0.12 & \\ \hline

\multicolumn{5}{c}{*p $\le$ 0.05, **p $\le$ 0.01, ***p $\le$ 0.001}
 \end{tabular}

\end{center}
\caption{This table shows
  results for regressions that combine H4N$\dnrm{RE}$
  with single-feature manual L\&M lists.}
\end{table}

%% file: appendix_tex/full_reg_4d_post_RE_neg_nocontr.tex
\begin{table}[h]
\begin{center}
  \small
\begin{tabular}{l|llll}
 var &  coeff &  std coeff &  t &  $R^2$\\ \hline\hline
 neg$\dnrm{lm}$  & -0.202**& -0.080& \textbf{-2.56}&1.02\\ \hline

 neg\dnrm{RE} &  -0.37*** &-0.111 & \textbf{-2.96}& 1.02\\ \hline

 neg\dnrm{ADD}&-0.033 &-0.0231& -1.03&1.00\\ \hline
 
neg$\dnrm{lm}$  & -0.0607& -0.0242& -0.38&1.02\\ 
neg\dnrm{RE} & -0.274 &-0.0822& -1.11& \\ \hline

neg\dnrm{RE} & -0.416*** &\textit{-0.124} & \textbf{-2.85}& 1.02\\ 
neg\dnrm{ADD}&\phantom{-} 0.0298 &\phantom{-} 0.0208& \phantom{-} 0.80 &\\ \hline

 neg$\dnrm{lm}$  & -0.0421& -0.0168& -0.27&1.02\\ 
neg\dnrm{RE} & -0.346 &-0.1037 & -1.35& \\ 
 neg\dnrm{ADD}&\phantom{-} 0.0277&\phantom{-} 0.0193& \phantom{-} 0.76&\\ \hline

\multicolumn{5}{c}{*p $\le$ 0.05, **p $\le$ 0.01, ***p $\le$ 0.001}
 \end{tabular}

\end{center}
\caption{This table shows
  results for regressions that combine RE, ADD and  L\&M dictionaries for the negative category.}
\end{table}

%% file: appendix_tex/full_reg_4d_post_RE_unc_nocontr.tex
\begin{table}[h]
\begin{center}
  \small
\begin{tabular}{l|llll}
 var &  coeff &  std coeff &  t &  $R^2$\\ \hline\hline
unc$\dnrm{lm}$  & -0.215*& -0.064& \textbf{-1.91}&1.01\\ \hline

unc\dnrm{RE} &  -0.377*** &-0.075 & \textbf{-2.77}& 1.02\\ \hline

unc\dnrm{ADD}&\phantom{-}0.0217&\phantom{-}0.0065& \phantom{-} 0.21&1.00\\ \hline
 
unc$\dnrm{lm}$  & \phantom{-}0.209& \phantom{-}0.0626& \phantom{-}0.45&1.01\\ 
unc\dnrm{RE} & -0.668 &\textit{-0.133}& -1.05& \\ \hline

unc\dnrm{RE} & -0.643*** &-0.128 & \textbf{-3.14}& 1.03\\ 
unc\dnrm{ADD}&\phantom{-}0.198 &\phantom{-}0.0594& \phantom{-}1.42 &\\ \hline

unc$\dnrm{lm}$  & -0.233& -0.0699& -0.42&1.03\\ 
unc\dnrm{RE} & -0.368 &-0.0736& -0.54& \\ 
unc\dnrm{ADD}&\phantom{-}0.234&\phantom{-}0.0702& \phantom{-}1.42&\\ \hline

\multicolumn{5}{c}{*p $\le$ 0.05, **p $\le$ 0.01, ***p $\le$ 0.001}
 \end{tabular}

\end{center}
\caption{This table shows
  results for regressions that combine RE, ADD and  L\&M dictionaries for the uncertain category.}
\end{table}

%% file: appendix_tex/full_reg_4d_post_RE_lit_nocontr.tex
\begin{table}[h]
\begin{center}
  \small
\begin{tabular}{l|llll}
 var &  coeff &  std coeff &  t &  $R^2$\\ \hline\hline
lit$\dnrm{lm}$  & -0.0291& -0.026& -0.83&1.00\\ \hline

lit\dnrm{RE} &  -0.056 &-0.028 & -0.55& 1.02\\ \hline

lit\dnrm{ADD}&-0.0195&-0.0156& -0.70&1.00\\ \hline
 
lit$\dnrm{lm}$  &-0.0759& \textit{-0.0683}& -0.95&1.00\\ 
lit\dnrm{RE} & \phantom{-}0.154&\phantom{-}0.077& \phantom{-}0.67& \\ \hline

lit\dnrm{RE} & -0.0261 &-0.0130& -0.20& 1.00\\ 
lit\dnrm{ADD}&-0.0136 &-0.0108& -0.39 &\\ \hline

lit$\dnrm{lm}$  & -0.0753& -0.0677& -0.94&1.00\\ 
lit\dnrm{RE} & \phantom{-} 0.155 &\phantom{-} 0.0775&\phantom{-} 0.66& \\ 
lit\dnrm{ADD}&-0.00107&-0.0008& -0.03&\\ \hline

\multicolumn{5}{c}{*p $\le$ 0.05, **p $\le$ 0.01, ***p $\le$ 0.001}
 \end{tabular}

\end{center}
\caption{This table shows
  results for regressions that combine RE, ADD and  L\&M dictionaries for the litigious category.}
\end{table}

%% file: appendix_tex/vol_HA_nocontr.tex
\begin{table}[h]
\begin{center}
  \small
\begin{tabular}{l|llll}
 var &  coeff &  std coeff &  t &  $R^2$\\ \hline\hline
  H4N\dnrm{RE} & \phantom{-}0.748*** &\phantom{-}\textit{0.224} & \phantom{-}\textbf{4.44}& 60.3\\ 
 neg$\dnrm{lm}$  & -0.096*& -0.038& \textbf{-2.55}&\\ \hline

 H4N\dnrm{RE}&\phantom{-} 0.741*** &\phantom{-}\textit{0.222}& \phantom{-}\textbf{4.30}&60.3\\ 
 all$\dnrm{lm}$   &-0.0438**&-0.0481&\textbf{-2.95} & \\ \hline
 H4N\dnrm{RE}&\phantom{-} 0.696*** &\phantom{-}\textit{0.208}& \phantom{-}\textbf{4.88}&60.3\\ 
 neg\_unc$\dnrm{lm}$   &-0.054&-0.032&-1.86& \\ \hline

 H4N\dnrm{RE}&\phantom{-} 0.693*** &\phantom{-}\textit{0.207}& \phantom{-}\textbf{4.24}&60.3\\ 
 neg\_lit$\dnrm{lm}$    &-0.034**&-0.037&\textbf{-2.70}& \\ \hline

\multicolumn{5}{c}{*p $\le$ 0.05, **p $\le$ 0.01, ***p $\le$ 0.001}
 \end{tabular}

\end{center}
\caption{This table shows
  results for regressions that combine H4N$\dnrm{RE}$
  with single-feature manual L\&M lists.}
\end{table}

%% file: appendix_tex/vol_neg_nocontr.tex
\begin{table}[h]
\begin{center}
  \small
\begin{tabular}{l|llll}
 var &  coeff &  std coeff &  t &  $R^2$\\ \hline\hline
 neg$\dnrm{lm}$  & \phantom{-}0.118***& \phantom{-}0.0472& \phantom{-}\textbf{3.30}&60.1\\ \hline

 neg\dnrm{RE} &  \phantom{-}0.219*** &\phantom{-}0.0657 & \phantom{-}\textbf{3.57}& 60.1\\ \hline

 neg\dnrm{ADD}&\phantom{-}0.032*** &\phantom{-}0.0224& \phantom{-}\textbf{4.06}&60.0\\ \hline

neg$\dnrm{lm}$  & \phantom{-}0.0014& \phantom{-}0.0005& \phantom{-}0.02&60.1\\ 
neg\dnrm{RE} & \phantom{-}0.217* &\phantom{-}0.065& \phantom{-}\textbf{1.96}& \\ \hline

neg\dnrm{RE} & \phantom{-}0.233** &\phantom{-}\textit{0.0699}& \phantom{-}\textbf{2.96}& 60.1\\ 
neg\dnrm{ADD}&-0.0087 &-0.006& -0.65 &\\ \hline

 neg$\dnrm{lm}$  & \phantom{-}0.00069& \phantom{-}0.0002& \phantom{-}0.01&60.1\\ 
neg\dnrm{RE} & \phantom{-}0.232* &\phantom{-}0.0696 & \phantom{-}\textbf{1.97}& \\ 
 neg\dnrm{ADD}&-0.0087&-0.006& -0.66&\\ \hline

\multicolumn{5}{c}{*p $\le$ 0.05, **p $\le$ 0.01, ***p $\le$ 0.001}
 \end{tabular}

\end{center}
\caption{This table shows
  results for regressions that combine RE, ADD and  L\&M dictionaries for the negative category.}
\end{table}

%% file: appendix_tex/vol_unc_nocontr.tex
\begin{table}[h]
\begin{center}
  \small
\begin{tabular}{l|llll}
 var &  coeff &  std coeff &  t &  $R^2$\\ \hline\hline
unc$\dnrm{lm}$  & \phantom{-}0.119*& \phantom{-}0.0356& \phantom{-}\textbf{2.25}&60.0\\ \hline

 unc\dnrm{RE} &\phantom{-}0.167*  &\phantom{-}0.0334&\phantom{-}\textbf{2.30}& 60.0\\ \hline

unc$\dnrm{ADD}$  &-0.013 &-0.0039& -0.17&  60.0\\ \hline

unc$\dnrm{lm}$  & \phantom{-}0.0432& \phantom{-}0.012& \phantom{-}0.28&60.0\\ 
unc\dnrm{RE} & \phantom{-}0.112 &\phantom{-}0.0224& \phantom{-}0.53& \\ \hline

unc\dnrm{RE} &\phantom{-}0.222***  &\phantom{-}0.0444&\phantom{-}\textbf{3.48}& 60.1\\ 
unc\dnrm{ADD}&-0.088 &-0.0263& -1.09 &\\ \hline

unc$\dnrm{lm}$  & \phantom{-}0.151& \phantom{-}\textit{0.0453}& \phantom{-}1.11&60.1\\ 
unc\dnrm{RE} & \phantom{-}0.0419 &\phantom{-}0.0083&\phantom{-}0.20& \\ 
 unc\dnrm{ADD}&-0.111&-0.0332&-1.41&\\ \hline

\multicolumn{5}{c}{*p $\le$ 0.05, **p $\le$ 0.01, ***p $\le$ 0.001}
 \end{tabular}

\end{center}
\caption{This table shows
  results for regressions that combine RE, ADD and  L\&M dictionaries for the uncertain category.}
\end{table}


%% file: appendix_tex/vol_lit_nocontr.tex
\begin{table}[h]
\begin{center}
  \small
\begin{tabular}{l|llll}
 var &  coeff &  std coeff &  t &  $R^2$\\ \hline\hline
lit$\dnrm{lm}$  & -0.0081& -0.0073& -0.62&60.0\\ \hline

 lit\dnrm{RE} &\phantom{-}0.0080  &\phantom{-}0.004&\phantom{-}0.20 & 60.0\\ \hline

 lit$\dnrm{ADD}$  & \phantom{-}0.028&\phantom{-}0.0224 & \phantom{-}1.07 &  60.0 \\ \hline

lit$\dnrm{lm}$  & -0.0635**& -0.057& \textbf{-2.93}&60.0\\ 
lit\dnrm{RE} & \phantom{-}0.181* &\phantom{-}\textit{0.0905}& \phantom{-}\textbf{2.46}& \\ \hline

lit\dnrm{RE} & -0.362 &-0.181 & -0.91& 60.0\\ 
lit\dnrm{ADD}&\phantom{-}0.041 &\phantom{-}0.0328&\phantom{-}1.50 &\\ \hline

 lit$\dnrm{lm}$  & -0.087***& -0.078&\textbf{-3.65}&60.1\\ 
lit\dnrm{RE} & \phantom{-}0.174* &\phantom{-}0.087& \phantom{-}\textbf{2.42}& \\ 
 lit\dnrm{ADD}&\phantom{-}0.066*&\phantom{-}0.0528& \phantom{-}\textbf{2.23}&\\ \hline

\multicolumn{5}{c}{*p $\le$ 0.05, **p $\le$ 0.01, ***p $\le$ 0.001}
 \end{tabular}

\end{center}
\caption{This table shows
  results for regressions that combine RE, ADD and  L\&M dictionaries for the litigious category.}
\end{table}